\documentclass[letterpaper]{article} 
\usepackage[draft]{aaai25}  
\usepackage{times}  
\usepackage{helvet}  
\usepackage{courier}  
\usepackage[hyphens]{url}  
\usepackage{graphicx} 
\usepackage{natbib}  
\usepackage{caption} 
\usepackage{algorithm}
\usepackage{algorithmic}
\usepackage{xcolor}
\usepackage{booktabs} 
\usepackage{multirow} 
\usepackage{newfloat}
\usepackage{listings}
\DeclareCaptionStyle{ruled}{labelfont=normalfont,labelsep=colon,strut=off} 
\floatstyle{ruled}
\newfloat{listing}{tb}{lst}{}
\floatname{listing}{Listing}
\title{Efficient Speech Command Recognition Leveraging Spiking Neural Network and Curriculum Learning-based Knowledge Distillation} 
\author{
    Jiaqi Wang\textsuperscript{\rm 1,2},
    Liutao Yu\textsuperscript{\rm 2},
    Liwei Huang\textsuperscript{\rm 2,3},
    Chenlin Zhou\textsuperscript{\rm 2},
    Han Zhang\textsuperscript{\rm 2,4},
    Zhenxi Song\textsuperscript{\rm 1},
    Min Zhang\textsuperscript{\rm 1},
    Zhengyu Ma\textsuperscript{\rm 2}\thanks{Corresponding author},
    Zhiguo Zhang\textsuperscript{\rm 1,2}\thanks{Corresponding author}
}
\affiliations{
    \textsuperscript{\rm 1}Harbin Institute of Technology Shenzhen,
    \textsuperscript{\rm 2} Peng Cheng Laboratory,
    \textsuperscript{\rm 3} Peking University,
    \textsuperscript{\rm 4} Harbin Institute of Technology

}

\usepackage{bibentry}

\begin{document}

\maketitle

\begin{abstract}
The intrinsic dynamics and event-driven nature of spiking neural networks (SNNs) make them excel in processing temporal information by naturally utilizing embedded time sequences as time steps. 
Recent studies adopting this approach have demonstrated SNNs' effectiveness in speech command recognition, achieving high performance by employing large time steps for long time sequences. However, the large time steps lead to increased deployment burdens for edge computing applications.
Thus, it is important to balance high performance and low energy consumption when detecting temporal patterns in edge devices. 
Our solution comprises two key components. 1). We propose a high-performance fully spike-driven framework termed SpikeSCR, characterized by a global-local hybrid structure for efficient representation learning, which exhibits long-term learning capabilities with extended time steps. 2). To further fully embrace low energy consumption, we propose an effective knowledge distillation method based on curriculum learning (KDCL), where valuable representations learned from the easy curriculum are progressively transferred to the hard curriculum with minor loss, striking a trade-off between power efficiency and high performance. We evaluate our method on three benchmark datasets: the Spiking Heidelberg Dataset (SHD), the Spiking Speech Commands (SSC), and the Google Speech Commands (GSC) V2. Our experimental results demonstrate that SpikeSCR outperforms current state-of-the-art (SOTA) methods across these three datasets with the same time steps. Furthermore, by executing KDCL, we reduce the number of time steps by 60\% and decrease energy consumption by 54.8\% while maintaining comparable performance to recent SOTA results. Therefore, this work offers valuable insights for tackling temporal processing challenges with long time sequences in edge neuromorphic computing systems.

\end{abstract}

%

\section{Introduction}
Recognized as the third generation of neural networks \cite{maass1997networks}, spiking neural networks (SNNs) effectively mimic the dynamics of biological neurons with sparse and asynchronous spikes. This approach enables SNNs to achieve high computational performance with low energy consumption, making them a viable energy-efficient alternative to artificial neural networks (ANNs) \cite{fang2023spikingjelly}.
Recent studies \cite{kugele2020efficient, yao2021temporal, kim2023exploring, zhu2024tcja} have demonstrated that the spiking mechanisms of SNNs excel in handling temporal data and event-driven applications, capitalizing on their inherent spatio-temporal encoding capabilities. 
These unique capabilities make SNNs exceptionally well-suited for processing constantly evolving dynamic content, such as speech and audio.  
While ANNs have demonstrated effectiveness in handling various acoustic features like Mel-frequency cepstrum coefficients \cite{logan2000mel}, Mel spectrograms \cite{shen2018natural}, and other audio representations in speech recognition tasks, their inherent architecture inevitably leads to high energy consumption \cite{yang2022lead}. This poses a non-negligible obstacle when deploying ANNs in speech command recognition (SCR) in energy-restricted scenarios, such as mobile and wearable devices, where both high accuracy and low energy consumption are crucial. In response, the natural temporal dynamics and event-driven property inherent in SNNs not only align with the properties of audio but also cater to the urgent needs for efficiency and effectiveness in speech processing systems, offering a promising solution for developing energy-efficient neuromorphic technologies \cite{xu2023reconfigurable}.

Since the inherent capabilities of SNNs align well with the requirements of SCR, an SNN-based model is essential to fully exploit these strengths. 
However, two critical challenges remain in developing such a high-performance and energy-efficient model for SCR. First, the efficient learning of contextual information across multi-level remains largely unexplored. While local context captures fine-grained details, it lacks an overall perspective, which will hinder the understanding of the command's intent and increase sensitivity to noise. Conversely, global context provides a broader understanding but may miss essential details needed to differentiate similar commands. Second, achieving a balance between high performance and energy efficiency has been largely overlooked. Recent SNN studies \cite{bittar2022surrogate,hammouamrilearning2024,deckers2024co} have leveraged the embedded time sequences as time steps. This approach aligns well with the dynamics of SNNs, as it avoids artificially increasing dimensions through repetition. However, directly using these time sequences can lead to higher accuracy but also increases the energy burden. A potential solution involves downsampling the time sequences to create shorter time steps \cite{wu2023dynamic,liu2024exploiting}, but this solution can compromise performance.


To address the aforementioned challenges in leveraging SNNs for SCR, our solution focuses on both enhancing the performance of SNNs and optimizing energy efficiency. 
Specifically, we introduce SpikeSCR, an innovative fully spike-driven framework. It is designed to capture both global and local contextual information within input sequences, thereby providing a fine-grained and comprehensive understanding. In terms of global representation learning, 
SpikeSCR develops a refined spike-driven attention mechanism, which builds on the spiking self attention (SSA) \cite{zhou2023spikformer} by integrating rotary position encoding (RoPE) \cite{su2024roformer}, leveraging the importance of positional information in temporal dynamics to improve contextual learning.
In terms of local representation learning, we propose a separable convolutional module equipped with a gated mechanism \cite{gulati2020conformer}, which efficiently captures critical dependencies by selectively controlling information flow, maintaining sparse computation and boosting performance with minimal resource usage.

Further, to address the high energy consumption issues associated with long time steps, we introduce a knowledge distillation method based on curriculum learning, termed KDCL. In KDCL, learning processes using long time steps are designated as easy curricula, whereas those involving shorter time steps are considered harder curricula. KDCL can efficiently and progressively transfer the valuable knowledge from models trained on longer time steps to those using shorter time steps with minor information loss, thus reducing energy consumption and maintaining performance levels comparable to SOTAs. This KDCL method has significant potential for improving performance in edge computation scenarios with limited time steps.

Our main contributions are summarized below:
\begin{itemize}
    \item We propose SpikeSCR, a fully spike-driven SNN framework with a global-local hybrid architecture for efficient representation learning and exhibiting long-term learning capabilities with extended time steps.

    \item We introduce KDCL, an effective curriculum learning-based knowledge distillation method, progressively transferring valuable representations from easy to hard curriculum with minor loss, achieving a satisfactory trade-off between energy consumption and performance.
    
    \item Extensive experiment results on SHD, SSC, and GSC datasets exhibit that SpikeSCR surpasses SOTAs with the same time steps. Through KDCL, it maintains comparable performance while reducing time steps by 60\% and energy consumption by 54.8\%.

\end{itemize}

\section{Related Works}
\subsection{Deep Spiking Neural Networks}
There are primarily two major approaches to obtain a trained SNN. One is ANNs-to-SNN conversion 
 \cite{deng2021optimal,bu2022optimal,xu2023constructing}, where a high-performance pre-trained ANN is transformed into an SNN by replacing its ReLU activation layers with spiking neurons \cite{rathi2021diet}.While the converted SNNs can achieve excellent performance, they often require a long-term simulation with hundreds or even thousands of time steps to accurately approximate the ReLU activations, which leads to large latency \cite{ su2023deep, guo2024ternary}. The other is direct training, the application of backpropagation-through-time (BPTT) on SNNs enables end-to-end training from scratch \cite{neftci2019surrogate}. Notably, this approach involves using surrogate gradient functions to approximate the derivatives of the non-differentiable Heaviside step function, which is essential for the spiking mechanism \cite{lee2016training, lee2020enabling}. Recent impressive works \cite{sengupta2019going, fang2021deep,zhou2023spikformer,yao2024spikedriven} have dedicated efforts to translating de-facto standard models from ANNs into new architectures that conform to the calculation characteristics of SNNs. 

\subsection{SNNs for Speech Command Recognition}
Speech command recognition, also known as keyword spotting, demands accurate recognition and low energy consumption, and it is crucial for numerous edge computing applications \cite{warden2018speech}. 
Recent SNNs have demonstrated considerable promise in meeting these requirements. 
Several studies \cite{yang2022deep,dampfhoffer2022investigating} have concentrated on enhancing network architecture to achieve high accuracy and energy efficiency in low-resource settings. Notably, further developments \cite{hammouamrilearning2024, sun2024ad, deckers2024co} involve novel methods for learning delays in SNNs, addressing the critical challenge of plastic delays for the timing of spike arrivals. Additionally, recent works \cite{bittar2022surrogate, zhang2024tc, deckers2024co} have proposed innovative spiking neurons for better learning dynamic time series. Moreover, substantial efforts \cite{stewart2023speech2spikes, cramer2020heidelberg} have been made to provide novel artificial cochleas models or spike generation pipelines, contributing to the advancement of neuromorphic speech recognition systems.

The aforementioned works often rely on long time sequences to approximate hundreds or more time steps, leading to a tendency towards higher energy consumption. The challenge of optimizing the balance between energy consumption and performance is unresolved. Resolving this challenge could significantly improve the efficiency and deployment of neuromorphic technologies in power-sensitive environments, enhancing both energy sustainability and the accessibility of speech processing applications.





\begin{figure*}[!ht]
\begin{center}
\includegraphics[width=0.97\linewidth]{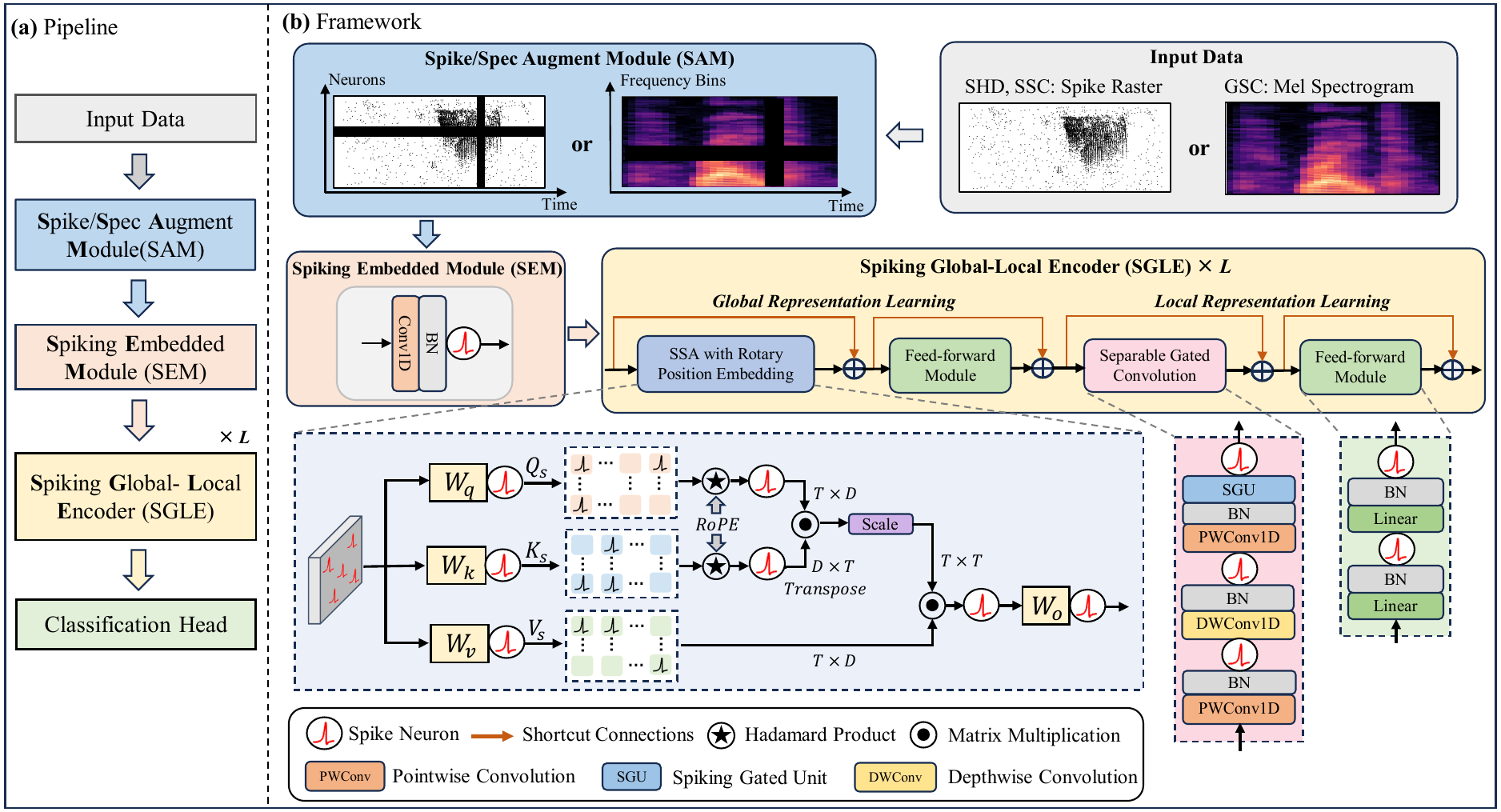}
\end{center}
\caption{Method Overview: (a) The pipeline of SpikeSCR. (b) The framework of SpikeSCR consists of a spike/spec augment module (SAM), a spiking embedded module (SEM), spiking global-local encoders (SGLE) with $L$ blocks, and a linear classification head. All the modules are spike-driven, consistent with SNN's computational characteristics.}
\label{fig:architecure}
\end{figure*}

\section{Methods}
\subsection{Spiking Neuron}
Spiking neurons are a crucial component of SNNs, which can serve as bio-plausible abstractions \cite{rathi2023exploring}. We adopt the Leaky Integrate-and-Fire (LIF) neuron, which has been widely proven effective \cite{Roy2019TowardsSM}. The dynamics of a LIF neuron are described as:
\begin{equation}\label{H[t]_LIF}
    H[t]=V[t-1]+\frac{1}{\tau}\left(X[t]-\left(V[t-1]-V_{reset}\right)\right),
\end{equation}
\begin{equation}\label{S[t]_LIF}
    S[t]=\Theta\left(H[t]-V_{th}\right),
\end{equation}
\begin{equation}\label{V[t]_LIF}
    V[t]=H[t]\left(1-S[t]\right)+V_{reset}S[t],
\end{equation}
where $\tau$ is the membrane time constant, $X[t]$ is the input current at time step $t$, $V_{reset}$ is the reset potential, $V_{th}$ is the firing threshold. Eq. (\ref{H[t]_LIF}) describes the update of membrane potential. Eq. (\ref{S[t]_LIF}) describes the spike generation process, where $\Theta(v)$ is the Heaviside step function: if $H[t] \geq V_{th}$ then $\Theta(v)=1$, meaning a spike is generated; otherwise $\Theta(v)=0$. $S[t]$ represents whether a neuron fires a spike at time step $t$. Eq. (\ref{V[t]_LIF}) describes the resetting process of membrane potential, where $H[t]$ and $V[t]$ represent the membrane potential before and after the evaluation of spike generation at time step $t$, respectively.

\subsection{SpikeSCR Architecture}

\subsubsection {Spike/Spec Augment Module (SAM).}
As shown in Figure \ref{fig:architecure}(b), we utilize the augmentation by SpecAugment \cite{park2019specaugment} with frequency mask and time mask for the Mel spectrogram of the input audio. For the spike trains, we utilize a method similar to EventDrop \cite{gu2021eventdrop}, which involves two operations: drop-by-time and drop-by-neuron, both performed at a fixed ratio to drop events in different dimensions randomly. Due to space constraints, the implementation details are provided in Appendix A.

\subsubsection {Spiking Embedded  Module (SEM).} The SEM is structured by sequentially stacking \{Conv1D-BN-LIF\}. Given the input \textbf{\textit{X}}, the process of SEM can be formulated as,
\begin{equation}
    SEM(\textbf{\textit{X}})=SN(BN(Conv(\textbf{\textit{X}}))),
\end{equation} 
where $SN$ is the spiking neuron. This structure confers several advantages. Firstly, the \{Conv1D-BN\}  ensures effective representation learning and sets up appropriate dimensions for the mechanism of multi-head attention computation in the subsequent SSA module. Secondly, the LIF neuron encodes the features into spike trains \cite{rathi2021diet, zhou2023spikformer, shen2024tim} as a spike generator.

\subsubsection {Spiking Global-Local Encoder (SGLE).}
Inspired by previous works \cite{vaswani2017attention, gulati2020conformer, yao2024spikedriven}, we introduce a novel architecture termed SGLE, designed to effectively capture both broad global and detailed local features, as illustrated in Figure \ref{fig:architecure}(b). 

For global representation learning, we utilize the SSA mechanism to capture long-range dependencies. To incorporate positional information in temporal dynamics, we added RoPE into the SSA. Given the spike input ${\textbf{\textit{X}}_S}$, the internal mechanisms are described as follows,
\begin{equation}
    Q_S=SN_Q(\textbf{\textit{W}}_Q(\textbf{\textit{X}}_S)),\;K_S=SN_K(\textbf{\textit{W}}_K(\textbf{\textit{X}}_S)),
\end{equation}
\begin{equation}
    Q_S'=SN_{R_Q}(R_Q\odot Q_S),\;K_S'=SN_{R_K}(R_K\odot K_S),
\end{equation}
\begin{equation}
    V_S=SN_V(\textbf{\textit{W}}_V(\textbf{\textit{X}}_S)),
\end{equation}
\begin{equation}
    SSA(Q_S',K_S',V_S) = SN\left((Q_S' K_S'^T V_S * s \right),
\end{equation}
where $SN$ is the spiking neuron, $\textbf{\textit{W}}$ is the learnable weight matrix of \{Linear-BN\}, $R$ is the rotary position matrix, $ \odot$ is the Hadamard Product, $s$ denotes a scale factor to avoid the gradient vanishing caused by the large integers. 

\begin{figure}[!t]
\begin{center}
\includegraphics[width=0.49\linewidth]{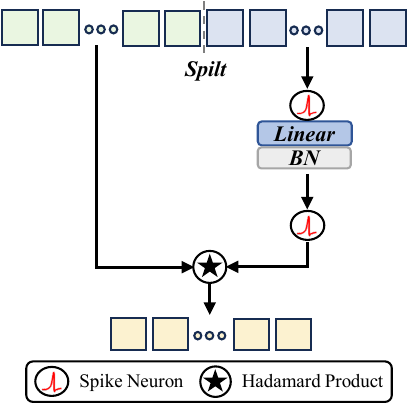}
\end{center}
\caption{Spiking Gated Unit.}\label{fig:sgu}
\end{figure}

For local representation learning, we first propose a spiking gated unit (SGU) module to enhance the performance of lightweight Conv-assemblies, as shown in Figure \ref{fig:sgu}. The SGU can be viewed as a spiking version of the Gated Linear Unit \cite{dauphin2017language, gulati2020conformer}, featuring a selective gating mechanism for information flow. It processes split feature embeddings $\textit{\textbf{X}}_1$ and $\textbf{\textit{X}}_2$, described as follows:
\begin{equation}
    SGU(\textbf{\textit{X}}_1,\textbf{\textit{X}}_2)=\textbf{\textit{X}}_2 \odot (SN_2(\textbf{\textit{W}}(SN_1(\textbf{\textit{X}}_1)))),
\end{equation}
where $\textbf{\textit{W}}$ is the learnable weight matrix of \{Linear-BN\}, $ \odot$ is the Hadamard Product. $SN_1$ aims to spikify the float-point inputs, $SN_2$ aims to regulate the transmission of information flow, and both are designed to implement sparse accumulate (AC) operations \cite{datta2022ace}. 
We then summarize the proposed separable gated convolution (SGC) into three blocks, operating on the input \textbf{\textit{X}} as follows:
\begin{equation}
    SGC(\textbf{\textit{X}})=SN(P_{WG}(SN(D_W(SN(P_{W}(\textbf{\textit{X}}))))))
\end{equation}
where $P_{W}$ is the pointwise (PW) convolution, $P_{WG}$ is the PW convolution with SGU, and $D_{W}$ is the depthwise convolution with a kernel size of 31. BN operations are considered foldable weights inserted into the convolutional stems.

\subsection{KDCL}

\subsubsection {Curriculum Learning.}
\begin{figure}[!t]
\begin{center}
\includegraphics[width=1.0\linewidth]{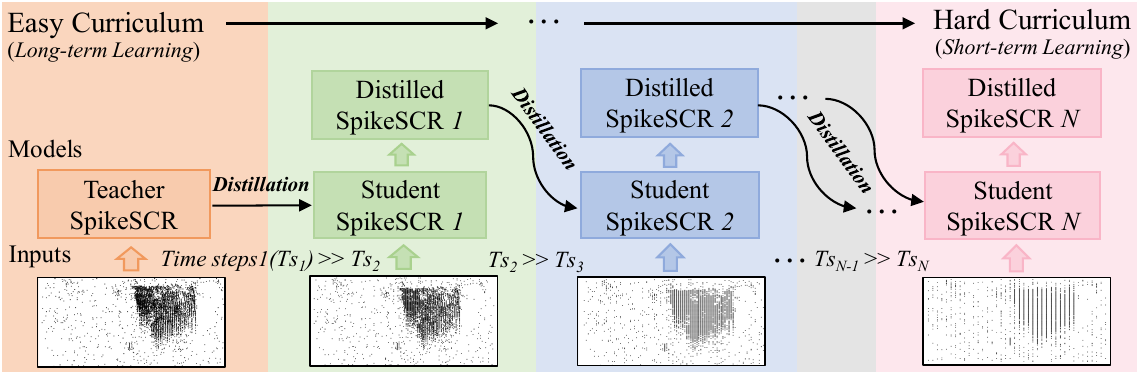}
\end{center}
\caption{Illustration of our proposed KDCL. Taking the "yes" command from the SSC dataset as an example, the KDCL starts with an easy curriculum to learn the first teacher model, and then progressively distills student models under the next curriculum setting until the hardest curriculum is completed. This process gradually enhances the model's performance from long time steps (long time sequences) to short time steps (short time sequences) through customized distillation across multiple curricula.}
\label{fig:KDCL}
\end{figure}

As illustrated in Figure \ref{fig:KDCL}, we propose an effective curriculum learning-based knowledge distillation method (KDCL) to optimize the trade-off between performance and energy consumption. We define the teacher model's initial learning from longer time steps as the easy curriculum. This knowledge is progressively transferred to student models with shorter time steps, defined as the hard curriculum. The learning process of KDCL is thoroughly described in Algorithm \ref{alg:knowledge_distillation}. This approach effectively enhances student model's performance by incrementally transferring valuable representations from long to short time steps.


\begin{algorithm}[!t]
\caption{KDCL}
\label{alg:knowledge_distillation}
\textbf{Require}: \\
\textit{Based on the easier curriculum}: A trained teacher SNN model \(T_1\); Training Dataset \(D_1\). \\
\textit{Based on the harder curriculum}: A new student SNN model \(S_1\); Training Dataset \(D_2\); \\
\textit{Multi-Task Loss Function} \(L_{MT}\): Eq. (\ref{Eq:multi}).


\textbf{Input}: \(T_1\), \(D_1\), \(S_1\), \(D_2\); Total training iteration \(I_{train}\); Target class label \(Y\). \\
\textbf{Output}: The well-trained SNN \(S_1\) \(\rightarrow\) The next curriculum level's teacher model \(T_2\).

\begin{algorithmic}[1] 
\STATE Frozen \(T_1\); Init \(S_1\) from \(T_1\); \(D_1\) and \(D_2\) synchronization.
\FOR{\(i = 1\) to \(I_{train}\)}
    \STATE Get mini-batch data \(x_1(i)\) from \(D_1\), mini-batch data \(x_2(i)\) from \(D_2\), and class label \(Y(i)\);
    \STATE Feed the \(x_1(i)\) into the \(T_1\), feed the \(x_2(i)\) into the \(S_1\);
    \STATE Calculate the \(T_1\) output \(O_{T_1}(i)\) and the \(S_1\) output \(O_{S_1}(i)\);
    \STATE Compute loss \(L_{MT}= \) \(\lambda_{1} \mathcal L_{CE}(O_{S_1}(i),Y(i)) + \) \\
     \(\lambda_{2} \mathcal L_{KD}(O_{T_1}(i),O_{S_1}(i))\);
    \STATE Calculate the derivative \(\nabla L_{MT}\) of the loss;
    \STATE Update the \(S_1\) weights \(W = W - \eta \nabla L_{MT}\) where \(\eta\) is learning rate.
\ENDFOR
\end{algorithmic}
\end{algorithm}

\subsubsection {Knowledge Distillation Method.} For clarity, we will describe the knowledge distillation process within a single curriculum.  It is straightforward to generalize this process to multiple curricula, since the same method is consistently applied in each curriculum setting. We denote $S(x)$ and $T(x)$ as the output logit vectors of the student and the teacher for input $x$, respectively. 
Formally, given a mini-batch, we first obtain the probability distribution $P_S(x;\tau)$ of the $S(x)$,
\begin{equation}
    P_S(x;\tau) = softmax\left({S(x)/ \tau}\right),
\end{equation}

where $\tau$ denotes a temperature parameter \cite{kim2021self}. The probability distribution $P_T(x;\tau)$ of the $T(x)$ is computed in the same manner. Further, we focus on the Kullback-Leibler (KL) divergence between the two probability distributions \cite{wang2022ltmd, yang2023snib}. The divergence $D_{KL}$ between $P_T(x;\tau)$ and $P_S(x;\tau)$ is computed for each instance as Eq. (\ref{DKL}). Finally, the weighted loss $L_{KD}$ is determined by averaging the $D_{KL}$ across all $M$ instances in the batch:
\begin{equation}\label{DKL}
D_{KL}(P_T || P_S) = \sum_i P_{T_i} \log \left( \frac{P_{T_i}}{P_{S_i}} \right),
\end{equation}
\begin{equation} \label{L_K}
\mathcal{L}_KD = \frac{1}{M} \sum_{i=1}^{M} D_{KL}(P_T \parallel P_S).
\end{equation}

\subsection {Training Strategy}
Note that the training of our model adopts the BPTT with surrogate gradients. We first introduce the input and output structure of SpikeSCR. Similar to the previous methods \cite{bittar2022surrogate,hammouamrilearning2024}, the proposed models take inputs of size $(B, T, N)$, where $B$ is the batch size, $T$ is the number of time steps, and $N$ is the number of input neurons or frequency bins. The number of classes is denoted by $Y$.  The classification head outputs are each neuron's synaptic potential through the time steps $T$. The method used to aggregate information from these time steps can be described as:
\begin{equation}
out_i[t] = softmax(s_i[t]) = \frac{e^{s_i[t]}}{\sum_{j=1}^{Y}e^{s_j[t]}}, 
\end{equation}
where $s_i[t](t=1,2,...,T)$ is the synaptic potential of neuron $i$ at time $t$. Then the final outputs of the model after $T$ time steps return the size $(B, C)$, which can be defined as: 
\begin{equation}
\hat{y}_i = \sum_{t=1}^{T} out_i[t].
\end{equation}

The cross-entropy loss $L_{CE}$ for the classification task is calculated as Eq. (\ref{Eq:CE}). For the multi-task in a specified curriculum, the training objective $L_{MT}$ is denoted as Eq. (\ref{Eq:multi}),
\begin{equation}\label{Eq:CE}
\mathcal{L}_{CE} = -\frac{1}{N} \sum_{n=1}^{N} \log(softmax(\hat{y}_n)[y_n]),
\end{equation}
\begin{equation}\label{Eq:multi}
\mathcal{L}_{MT} = \lambda_{1} \mathcal{L}_{CE} + \lambda_{2} \mathcal{L}_{KD},
\end{equation}
where $L_{KD}$ is detailed in Eq. (\ref{L_K}). We set $\lambda_{1}=1.0$ and $\lambda_{2}=0.5$, which is refined through experiments, to balance the gradients in the overall learning process.

\section{Experiments}
\subsection{Datasets and Experimental Setup}
We evaluate our methods on two spiking datasets, SHD (Spiking Heidelberg Digits) and SSC (Spiking Speech Commands) \cite{cramer2020heidelberg}, and a non-spiking counterpart of SSC, GSC (Google Speech Commands V2) \cite{warden2018speech}. Both spiking datasets are constructed using artificial cochlear models to convert audio recording data to spikes. The SHD dataset consists of 10k recordings of 20 different classes, with spoken digits ranging from zero to nine in both English and German. SSC and GSC are much larger datasets, each containing 100k recordings. The task on SSC and GSC is the classification of all 35 different speech commands. We utilize the same preprocess methods \cite{hammouamrilearning2024} to all three datasets. For the two spiking datasets, the input neurons were reduced from 700 to 140 by using spatio-temporal bins across every 5 neurons. As for the temporal dimension,
we used a discrete time-step (which means at each $\Delta t$ interval, the number of spikes is extracted) and a zero right-padding to ensure the same time duration (used as the time steps) of all recordings in one batch. For the GSC dataset, we used the Mel spectrogram of the waveforms with 140 frequency bins and approximately the same time steps consistent with the SSC. The dataset statistics and experimental setups are detailed in Appendix B. The empirical curriculum setting in KDCL is described as follows. For the SHD dataset, KDCL training involves two curricula from easy to hard: 100 time steps and 40 time steps, as the smaller dataset risks overfitting with excessive time steps. For the larger SSC and GSC datasets, KDCL involves four curricula from easy to hard: 500 (400), 200, 100, and 40 time steps.

\begin{table*}[!t]
\centering
\resizebox{\textwidth}{!}{%
\begin{tabular}{@{}clccc@{}}
\toprule
\textbf{Dataset} & \textbf{Model} & \textbf{Param (M)} & \textbf{Time Steps} & \textbf{Acc (\%)} \\ \midrule
\multirow{7}{*}{SHD}
                     & SpikeGRU \cite{dampfhoffer2022investigating} & --- & 250 & 87.80 \\
                     & TC-LIF \cite{zhang2024tc} & 0.19 & --- & 88.91 \\
                     & Adaptive Axonal Delays \cite{sun2024ad} & 0.11 & 150 & 92.45 \\
                     & DL256-SNN-Dloss \cite{sun2023learnable} & 0.21 & 250 & 93.55 \\
                     & \textbf{SpikeSCR (1L-8-128)} & \textbf{0.25} & \textbf{40} & \textbf{93.60\textsuperscript{\textdagger}/92.01} \\
                     & RadLIF \cite{bittar2022surrogate} & 3.90 & 100 & 94.62 \\
                     & DCLS-Delays (2L-1KC) \cite{hammouamrilearning2024} & 0.20 & 100 & 95.07 \\
                     & \textbf{SpikeSCR (1L-8-128)} & \textbf{0.25} & \textbf{100} & \textbf{95.70} \\
\midrule 
\multirow{10}{*}{SSC} & TC-LIF \cite{zhang2024tc} & 0.14 & --- & 61.09 \\
                        & SpikeGRU \cite{dampfhoffer2022investigating} & --- & 250 & 77.00 \\
                        & RadLIF \cite{bittar2022surrogate} & 3.90 & 100 & 77.40 \\
                        & DCLS-Delays (2L-2KC) \cite{hammouamrilearning2024} & 1.40 & 100 & 80.16 \\
                        & d-cAdLIF \cite{deckers2024co} & 0.70 & 100 & 80.23 \\
                        & DCLS-Delays (3L-2KC) \cite{hammouamrilearning2024} & 2.50 & 100 & 80.69 \\
                        & \textbf{SpikeSCR (1L-16-256)} & \textbf{1.63} & \textbf{40} & \textbf{79.76\textsuperscript{\textdagger}/79.12} \\
                        & \textbf{SpikeSCR (1L-16-256)} & \textbf{1.63} & \textbf{100} & \textbf{82.99\textsuperscript{\textdagger}/82.54} \\
                        & \textbf{SpikeSCR (2L-16-256)} & \textbf{3.15} & \textbf{40} & \textbf{80.25\textsuperscript{\textdagger}/79.23} \\
                        & \textbf{SpikeSCR (2L-16-256)} & \textbf{3.15} & \textbf{100} & \textbf{83.69\textsuperscript{\textdagger}/82.79} \\                        
\midrule 
\multirow{10}{*}{GSC} & Speech2Spike \cite{stewart2023speech2spikes} & --- & 200 & 88.50 \\
                         & TC-LIF \cite{zhang2024tc} & 0.61 & --- & 88.91 \\
                         & RadLIF \cite{bittar2022surrogate} & 1.20 & 100 & 94.51 \\
                         & DCLS-Delays (2L-2KC) \cite{hammouamrilearning2024} & 1.40 & 100 & 95.00 \\
                         & Fully Spiking LMU Blocks \cite{liulmuformer} & 1.62 & --- & 95.20 \\
                         & DCLS-Delays (3L-2KC) \cite{hammouamrilearning2024} & 2.50 & 100 & 95.35 \\
                         & \textbf{SpikeSCR (1L-16-256)} & \textbf{1.63} & \textbf{40} & \textbf{94.71\textsuperscript{\textdagger}/94.50} \\
                         & \textbf{SpikeSCR (1L-16-256)} & \textbf{1.63} & \textbf{100} & \textbf{95.90\textsuperscript{\textdagger}/95.56} \\
                         & \textbf{SpikeSCR (2L-16-256)} & \textbf{3.15} & \textbf{40} & \textbf{95.01\textsuperscript{\textdagger}/94.54} \\ 
                         & \textbf{SpikeSCR (2L-16-256)} & \textbf{3.15} & \textbf{100} & \textbf{96.08\textsuperscript{\textdagger}/95.60} \\ 
                         \bottomrule
\end{tabular}
}
\caption{Comparison of model performances with prior SNN works on three different datasets, SHD, SSC and GSC. \textdagger indicates the performance after KDCL. The notation of (nL-m-d) in the table specifies the model architecture, where n represents the number of SGLE blocks, m represents the number of attention heads, and d represents the hidden size. For DCLS (nL-mKC), nL represents the number of layers and mKC is the number of kernel counts in dilated convolutions. }
\label{main-resuts}
\end{table*}

\subsection{Main Results}

We compare our method to recent SNN works on the SHD (20 classes), SSC, and GSC (both with 35 classes) benchmark datasets in terms of accuracy and model size, as shown in Table \ref{main-resuts}. 
Even without knowledge transfer, our models surpass recent SOTA approaches across three benchmarks with the same time steps, including those focusing on learning delays in SNNs and modeling neuron dynamics. Notably, SpikeSCR achieves an accuracy of 82.54\% on the SSC dataset with 100 time steps, demonstrating significant improvements compared to the current method DCLS \cite{hammouamrilearning2024}, which achieves 80.69\% accuracy with more parameters (2.5M vs. 1.63M). Moreover, the application of KDCL results in substantial improvements, achieving competitive results of 93.60\%, 80.25\%, and 95.01\% on the SHD, SSC, and GSC datasets, respectively, with the fewest time steps (40) among compared works. 
Comprehensive results across all datasets are detailed in Appendix C due to space constraints.

\begin{figure*}[!t]
\begin{center}
\includegraphics[width=1.0\linewidth]{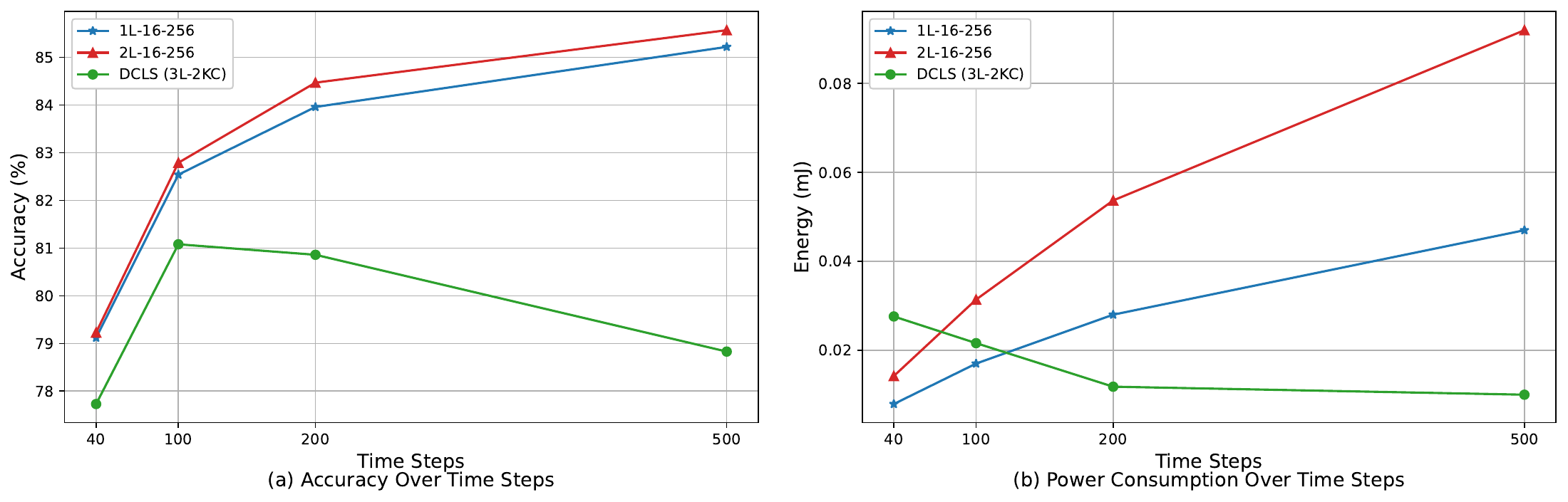}
\end{center}
\caption{Comparative analysis of model performance and energy consumption from short to long time steps on SSC dataset.}\label{fig:trade-off compare_ssc}
\end{figure*}

We further validate our models' capability for long-term learning over extended time steps, which is crucial for demonstrating the effectiveness of using a model with large time steps (such as 500 time steps) as the initial teacher model in subsequent KDCL procedures.
As shown in Figure \ref{fig:trade-off compare_ssc}(a), we present the results of the SpikeSCR trained with either a single block or two blocks of SGLE from scratch on the SSC dataset with various time steps: 40, 100, 200 and 500.  We also compare our results with the reproduced DCLS method. The accuracy of SpikeSCR clearly improves as the number of time steps increases, rising from 79.23\% to 85.57\%. In contrast, DCLS shows optimal performance with 100 time steps but then declines with more time steps. This comparison indicates that SpikeSCR is better suited for long-term learning, effectively leveraging both comprehensive global contexts and detail-oriented local contexts. Figure \ref{fig:trade-off compare_ssc}(b) further illustrates the energy consumption results. As the number of time steps decreases, the energy consumption of SpikeSCR declines significantly. Specifically, from 100 to 40 time steps, the energy consumption for one block (blue) decreases by 53.8\% (from 0.0171mJ to 0.0079mJ), and for two blocks (red), it decreases by 54.8\% (from 0.0314mJ to 0.0142mJ). Notably, SpikeSCR demonstrates a significant energy efficiency advantage with 40 time steps, compared to the DCLS model.

\begin{figure}[!t]
\begin{center}
\includegraphics[width=1.0\linewidth]{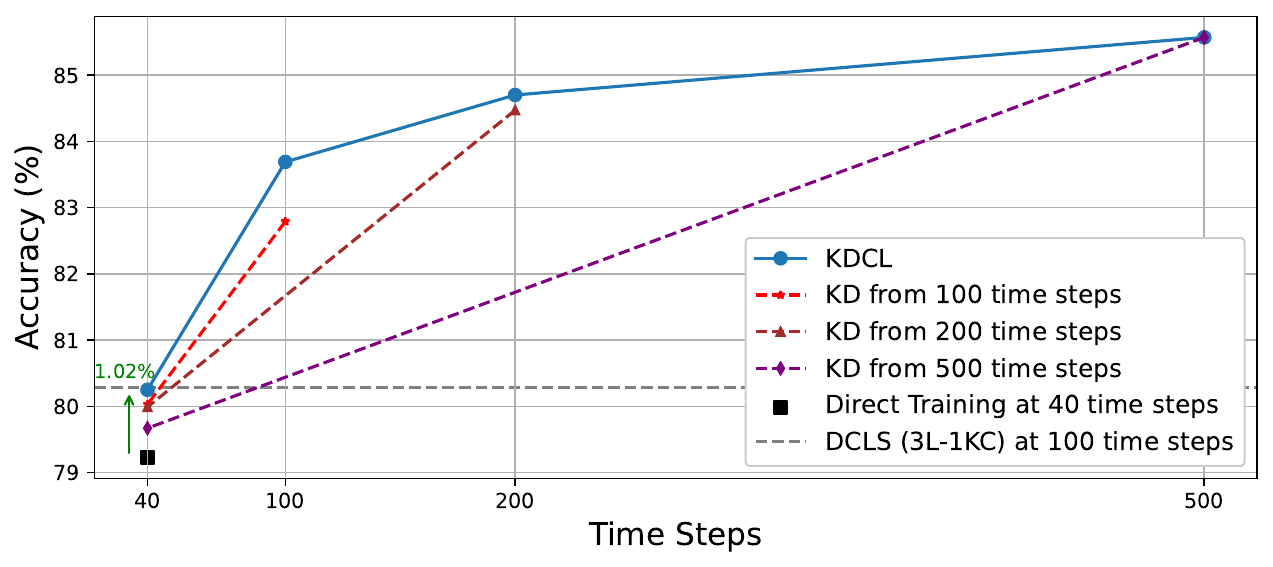}
\end{center}
\caption{Comparing the performance of direct training, knowledge distillation, and KDCL on SSC dataset.}\label{fig:kdcl_acc_ssc_2l}
\end{figure}

As in Figure \ref{fig:kdcl_acc_ssc_2l}, KDCL-based models demonstrate superior performance compared to those trained with the same number of time steps under the two blocks of SGLE. The models trained with KDCL consistently outperform those trained directly with any specified number of time steps. Notably, while models trained with large time steps (e.g., 500) achieve high performance, directly knowledge distillation from these teacher models to student models with merely 40 time steps yields marginal improvements (80.25\% vs. 80.03\%, 79.99\%, and 79.67\%). This suggests that excessively transferring knowledge from models with longer time steps to models with shorter time steps can be challenging in both performance and convergence. KDCL leads to the best results at 40 time steps, improving accuracy by 1.02\% compared to direct training (80.25\% vs. 79.23\%). The comparable performance with SOTA methods like DCLS (2L-2KC) (80.25\% vs. 80.16\%; 40 steps vs. 100 steps) highlights its effectiveness with limited time steps.



\begin{figure}[!t]
\begin{center}
\includegraphics[width=1.0\linewidth]{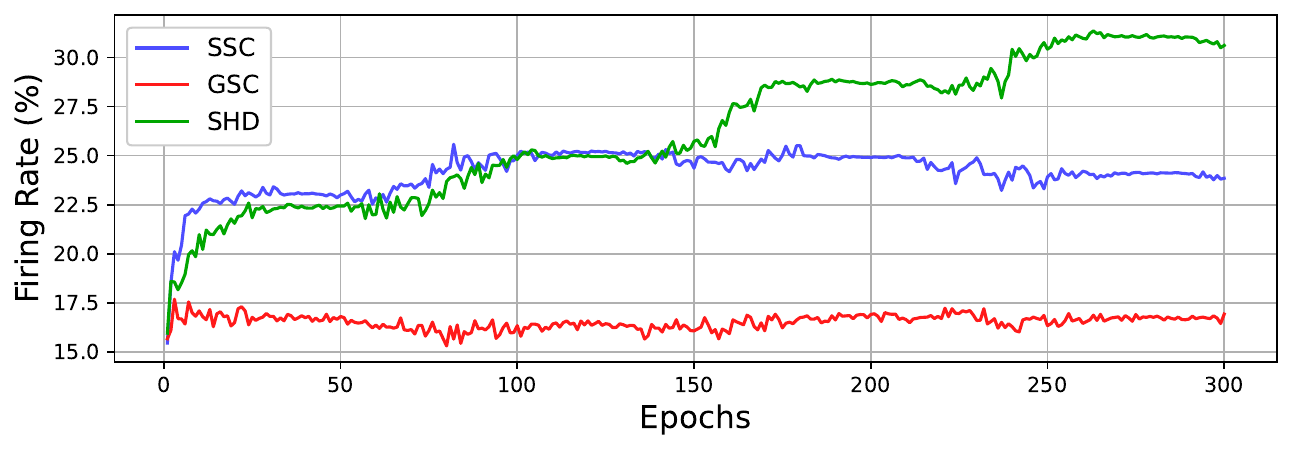}
\end{center}
\caption{Firing rate of the LIF Neuron $SN_2$ in SGU on SHD, SSC and GSC datasets.}
\label{fig:fire_rate}
\end{figure}

Figure \ref{fig:fire_rate} illustrates the firing rate trends during the training process of the $SN_2$ neuron within the proposed SGU module across three datasets: SHD, SSC, and GSC.  Over 300 training epochs with 100 time steps in a single block of SGLE, the firing rate initially increases and then stabilizes in the final 50 epochs, reaching approximately 30.6\% for SHD, 23.8\% for SSC, and 16.9\% for GSC. This stability and variability suggest that the SGU can adapt to different data patterns while efficiently maintaining gating mechanisms within the information flow and avoiding multiply-accumulate (MAC) operations \cite{chen2023training}, thereby validating the rationale and effectiveness of the design.


Table \ref{fig:compare_pe} presents the impact of different plug-and-play positional encodings on model performance and energy consumption. Compared to the model without PE, the sinusoidal positional encoding (SinPE) from the vanilla Transformer \cite{vaswani2017attention} improves performance by 1.24\%. However, SinPE also increases energy consumption by nearly 5.2 times, as it transforms spike-form data into float-form data with positional information, thus increasing MAC operations. A LIF neuron layer integrated after applying SinPE could preserve the positional information in the spike sequences to keep low energy consumption, but only achieves 85.58\% accuracy. In contrast, the integration of RoPE and LIF within the SSA maintains energy consumption levels comparable to those without positional encoding, while achieving the highest accuracy of 95.70\%.

Comprehensive ablation studies are conducted with one SGLE block on the GSC dataset under 100 time steps, summarized in Table \ref{fig:ablation}. Various components are systematically removed from the model, leading to performance reductions. The removal of the KDCL method results in a 0.34\% decrease in accuracy. Further, eliminating the augment module reduces accuracy by an additional 0.57\%. The impact of removing RoPE from the global representation learning module leads to a 0.34\% reduction in accuracy. Omitting the spiking gated unit from the local representation learning module results in a 0.8\% decrease. Moreover, the absence of the separable convolution module ($P_{W}+D_{W}+P_{W}$) causes a more significant decline, reducing accuracy by 1.08\%.

\begin{table}[!t]
\centering
\begin{tabular}{@{}clcc@{}}
\toprule
\textbf{Dataset} & \textbf{Position Encoding} & \textbf{Energy (mJ)} & \textbf{Acc (\%)} \\ \midrule
 \multirow{4}{*}{SHD}
 & None & 0.005 &  93.69 \\
 & Sinusoidal PE (SinPE)  & 0.026 & 94.93 \\
 & SinPE + LIF & 0.004 & 85.58 \\
 & RoPE + LIF & 0.005 & 95.70 \\ 
\bottomrule
\end{tabular}
\caption{The impact of different positional encodings on model performance and energy consumption.}
\label{fig:compare_pe}
\end{table}

\begin{table}[!t]
\resizebox{0.495\textwidth}{!}{%
\centering
\begin{tabular}{@{}clcc@{}}
\toprule
\textbf{Dataset} & \textbf{Model} & \textbf{Param (M)} & \textbf{Acc (\%)} \\ \midrule
\multirow{6}{*}{GSC} & SpikeSCR (1L-16-256) & 1.63 & 95.90 \\
                         & w/o KDCL & 1.63 & 95.56 \\
                         & w/o SAM & 1.63 & 94.99 \\
                         & w/o RoPE & 1.63 & 94.65 \\
                         & w/o SGU & 1.50 & 93.85 \\
                         & w/o Separable Convolution  & 0.87 & 92.77 \\ 
                         \bottomrule
\end{tabular}
}
\caption{Ablation studies on GSC dataset (100 time steps).}
\label{fig:ablation}
\end{table}


\section{Conclusion}
In this study, we first propose a fully spike-driven framework, SpikeSCR, aiming at enhancing speech command recognition. SpikeSCR effectively integrates position-embedded SSA and separable gated convolution to achieve global and local contextual representation learning. We demonstrate its efficient long-term learning capabilities, achieving excellent performance with extended time steps. To achieve a satisfactory trade-off between energy consumption and performance, we further introduce an effective knowledge distillation method, KDCL, where valuable representations learned from easy curriculum are progressively transferred to hard curriculum with minor loss. Our experimental results show that SpikeSCR outperforms SOTA methods across three benchmark datasets, achieving superior performance at the same time steps. Furthermore, through KDCL, SpikeSCR maintains comparable performance to SOTA results while reducing time steps by 60\% and energy consumption by 54.8\%.



\clearpage
\bibliography{aaai25}

\clearpage
\appendix
\section{Appendix}

\subsection{Appendix A: Augments}\label{appendixa}
\begin{figure}[!htbp]
\begin{center}
\includegraphics[width=1.0\linewidth]{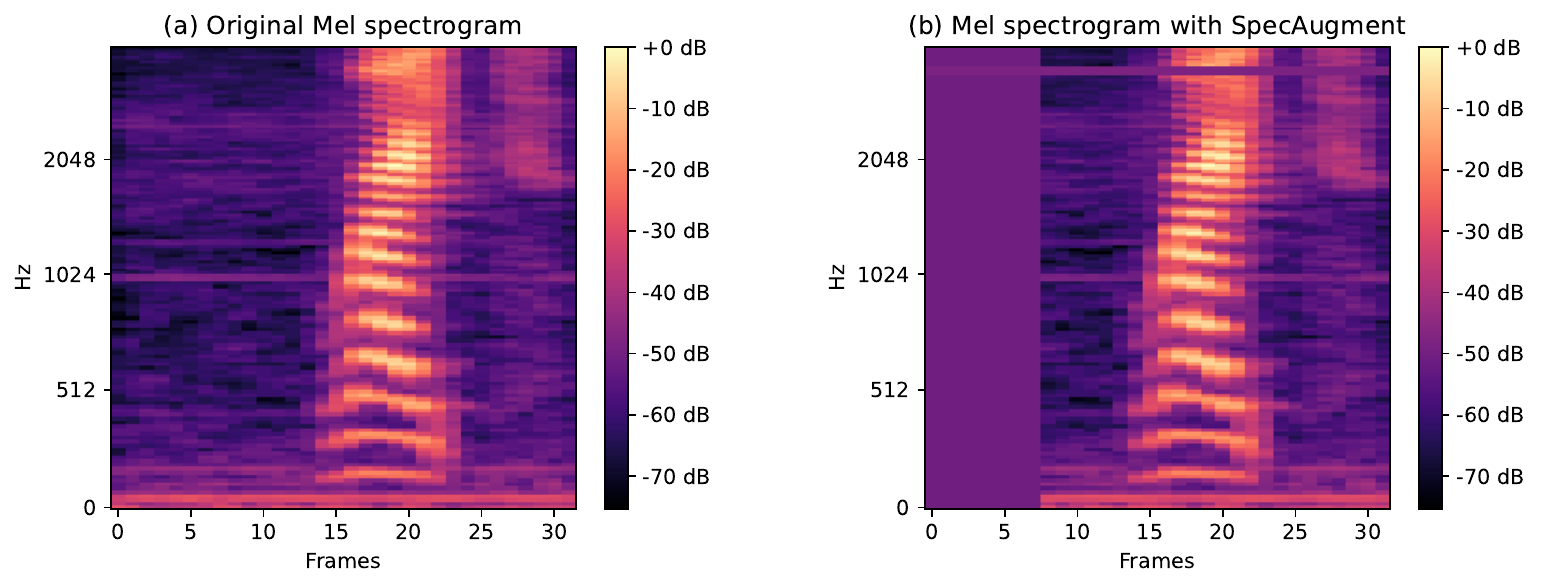}
\end{center}
\caption{Comparison of the "Right" command in GSC dataset before and after augmentation.}\label{fig:gsc_combined_5}
\end{figure}

\begin{figure}[!htp]
\begin{center}
\includegraphics[width=1.0\linewidth]{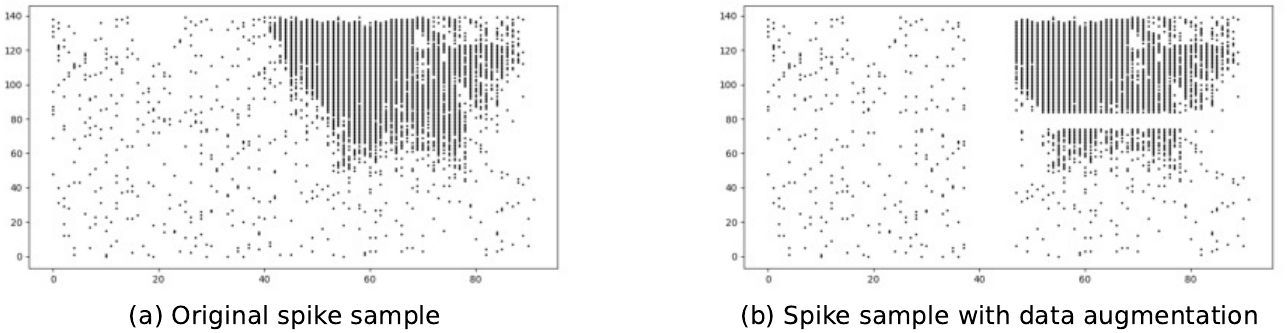}
\end{center}
\caption{Comparison of the "Cat" command in SSC dataset before and after augmentation.}\label{fig:ssc_combined_61}
\end{figure}

We utilize SpecAugment \cite{park2019specaugment} with frequency masking and time masking for the Mel spectrogram of the input audio. The specific parameter settings for SpecAugment are detailed in Table \ref{tab:augforgscshdssc}. Additionally, we present a comparison of the augmented data in Figure \ref{fig:gsc_combined_5}, illustrating a "Right" command sample before and after data augmentation. For the spike trains in SHD and SSC, we utilize a method similar to EventDrop \cite{gu2021eventdrop} involves two operations: drop-by-time and drop-by-neuron, both performed at a fixed ratio to drop events in different dimensions randomly. The specific parameter settings for EventDrop are detailed in Table \ref{tab:augforgscshdssc}. We also present a comparison of the augmented data in Figure \ref{fig:ssc_combined_61}, illustrating a "Cat" command sample before and after data augmentation.

\begin{table}[!htbp]
\centering
\begin{tabular}{@{}lc@{}}
\toprule
\textbf{Parameter} & \textbf{GSC} \\ 
\midrule
Number of Frequency Masks & 1 \\
Frequency Mask Size & 10 \\
Number of Time Masks & 1 \\
Time Mask Size & 0.25 \\
\midrule
\multicolumn{1}{l}{\textbf{Parameter}} & \multicolumn{1}{r}{\textbf{SHD / SSC}} \\
\midrule
Drop Proportion & 0.5 / 0.5 \\
Time Drop Size & 0.2 / 0.1 \\
Neuron Drop Size & 20 / 10 \\
\bottomrule
\end{tabular}
\caption{Augmentation parameters for three datasets.}
\label{tab:augforgscshdssc}
\end{table}

\subsection{Appendix B: Datasets and Experimental Setup}\label{appendixb}
The dataset statistics of three benchmark datasets are shown in Table \ref{datasets}. The SHD dataset consists of recordings of spoken digits ranging from zero to nine in English and German language. There are only training and testing sets, with a total of 10,420 samples in SHD dataset. SSC is the spiking version of GSC. Both SSC and GSC have the same total number of samples, 105,829, which consist of a total of 35 word commands (Yes, No, Up, Down, Left, Right, etc), although the number of training and testing samples differ slightly. Overall, the number of samples of SSC and GSC are nearly ten times that of SHD. Therefore, we set two learning curricula for SHD and four learning curricula for SSC and GSC. Our work is implemented using the PyTorch-based SpikingJelly \cite{fang2023spikingjelly} framework. Additionally, we provide detailed experimental settings for the three datasets, as shown in Table \ref{tab:training_config}. It is worth noting that to ensure the stability of the knowledge distillation process, we further employ a warming-up strategy, setting it for the initial 10 epochs.
\begin{table}[!htbp]
\centering
\begin{tabular}{@{}clll@{}}
\toprule
Datasets &
  \begin{tabular}[c]{@{}c@{}}Training \\Sample\end{tabular} &
  \begin{tabular}[c]{@{}c@{}}Validation\\Sample\end{tabular} &
  \begin{tabular}[c]{@{}c@{}}Testing \\Sample\end{tabular} \\ \midrule
    SHD & 8156   & ——   & 2264  \\
    SSC & 75466  & 9981 & 20382 \\
    GSC & 84843 & 9981 & 11005 \\ \bottomrule
\end{tabular}
\caption{Dataset statistics of three datasets.}
\label{datasets}
\end{table}

\begin{table*}[!htbp]
\centering
\begin{tabular}{@{}lp{3cm}p{3cm}p{3cm}@{}}
\toprule
Datasets & SHD & SSC & GSC \\ \midrule
Direct Training Epochs & 300 & 300 & 300 \\
Knowledge Distillation Epochs & 300 & 400 & 400 \\
Batch Size & \multicolumn{3}{c}{256} \\
Optimizer & \multicolumn{3}{c}{AdamW} \\
Weight Decay & 1e-2 & 1e-2 & 5e-3 \\
Surrogate Function & \multicolumn{3}{c}{Atan ($\alpha$=5.0)} \\
Learning Rate & 1e-2 & 5e-3 & 2e-3 \\
LR Scheduler & \multicolumn{3}{c}{Cosine Annealing, $T_{\max}$=40} \\
Spiking Neuron & \multicolumn{3}{c}{LIF ($\tau$=2.0, $V_{threshold}$=1.0)} \\
GPU & \multicolumn{3}{c}{RTX 4090} \\ \bottomrule
\end{tabular}
\caption{Implementation details for SHD, SSC, and GSC datasets.}
\label{tab:training_config}
\end{table*}

\begin{figure}[!b]
\begin{center}
\includegraphics[width=1.0\linewidth]{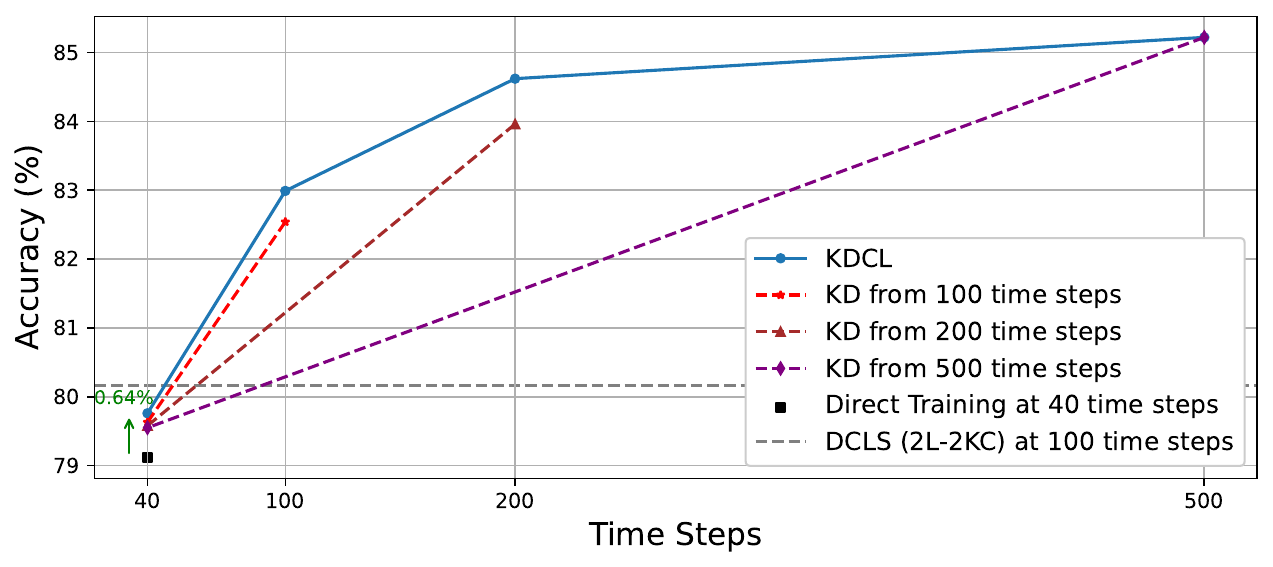}
\end{center}
\caption{Comparing the performance of SpikeSCR with a single block of SGLE among direct training, knowledge distillation, and KDCL on SSC dataset.}\label{fig:kdcl_acc_ssc_1l}
\end{figure}

\begin{table*}[!ht]
\centering
\begin{tabular}{@{}clcccc@{}}
\toprule
\textbf{Dataset} & \textbf{Model}  & \textbf{Time Steps} & \textbf{Param (M)} & \textbf{Energy (mJ)} & \textbf{Acc (\%)} \\ 
\midrule
\multirow{12}{*}{SSC}     & DCLS-Delays (3L-2KC)& 40  & 2.34 & 0.0276 & 77.73 \\
                             & SpikeSCR (1L-16-256)     & 40 & 1.63 & 0.0079 & 79.12 \\
                             & SpikeSCR (2L-16-256)     & 40 & 3.15 & 0.0142 & 79.23 \\
\cmidrule(l{4pt}r{4pt}){2-6} 
                             & DCLS-Delays (3L-2KC) & 100 & 2.34 & 0.0216 & 81.08 \\
                             & SpikeSCR (1L-16-256)     & 100 & 1.63 & 0.0171 & 82.54 \\
                             & SpikeSCR (2L-16-256)     & 100 & 3.15 & 0.0314 & 82.79 \\
\cmidrule(l{4pt}r{4pt}){2-6} 
                             & DCLS-Delays (3L-2KC) & 200 & 2.34 & 0.0118 & 80.86 \\
                             & SpikeSCR (1L-16-256)     & 200 & 1.63 & 0.0270 & 83.96 \\
                             & SpikeSCR (2L-16-256)     & 200 & 3.15 & 0.0537 & 84.47 \\
\cmidrule(l{4pt}r{4pt}){2-6} 
                             & DCLS-Delays (3L-2KC) & 500 & 2.34 & 0.0110 & 78.76 \\
                             & SpikeSCR (1L-16-256)     & 500 & 1.63 & 0.0472 & 85.22 \\
                             & SpikeSCR (2L-16-256)     & 500 & 3.15 & 0.0920 & 85.57 \\ 
\bottomrule
\end{tabular}
\caption{Quantitative comparison of performance and energy consumption between SpikeSCR and DCLS across various time steps on SSC dataset.}
\label{table:performance_ssc}
\end{table*}

\begin{figure*}[!t]
\begin{center}
\includegraphics[width=1.0\linewidth]{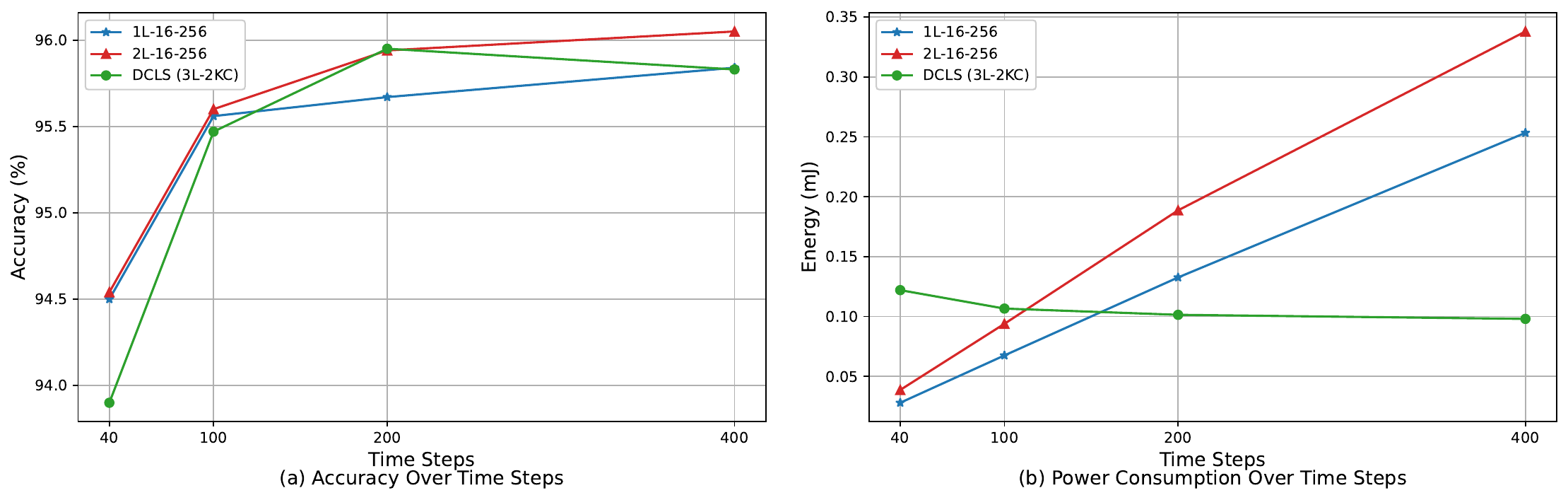}
\end{center}
\caption{Comparative analysis of performance and energy consumption from short to long time steps on GSC dataset.}\label{fig:trade-off compare_gsc}
\end{figure*}

\begin{table*}[!ht]
\centering
\begin{tabular}{@{}clcccc@{}}
\toprule
\textbf{Dataset} & \textbf{Model} & \textbf{Time Steps} & \textbf{Param (M)} & \textbf{Energy (mJ)} & \textbf{Acc (\%)} \\ 
\midrule
\multirow{12}{*}{GSC}        & DCLS-Delays (3L-2KC) & 40 & 2.34 & 0.1220 & 93.90 \\
                             & SpikeSCR (1L-16-256)     & 40 & 1.63 & 0.0280 & 94.50 \\
                             & SpikeSCR (2L-16-256)     & 40 & 3.15 & 0.0387 & 94.54 \\
\cmidrule(l{4pt}r{4pt}){2-6} 
                             & DCLS-Delays (3L-2KC) & 100 & 2.34 & 0.1066 & 95.47 \\
                             & SpikeSCR (1L-16-256)     & 100 & 1.63 & 0.0675 & 95.56 \\
                             & SpikeSCR (2L-16-256)     & 100 & 3.15 & 0.0939 & 95.60 \\
\cmidrule(l{4pt}r{4pt}){2-6} 
                             & DCLS-Delays (3L-2KC) & 200 & 2.34 & 0.1014 & 95.95 \\
                             & SpikeSCR (1L-16-256)     & 200 & 1.63 & 0.1326 & 95.67 \\
                             & SpikeSCR (2L-16-256)     & 200 & 3.15 & 0.1886 & 95.94 \\
\cmidrule(l{4pt}r{4pt}){2-6} 
                             & DCLS-Delays (3L-2KC) & 400 & 2.34 & 0.0980 & 95.83 \\
                             & SpikeSCR (1L-16-256)     & 400 & 1.63 & 0.2533 & 95.84 \\
                             & SpikeSCR (2L-16-256)     & 400 & 3.15 & 0.3379 & 96.05 \\ 
\bottomrule
\end{tabular}
\caption{Quantitative comparison of performance and energy consumption between SpikeSCR and DCLS across various time steps on GSC dataset.}
\label{table:performance_gsc}
\end{table*}

\begin{figure}[!t]
\begin{center}
\includegraphics[width=1.0\linewidth]{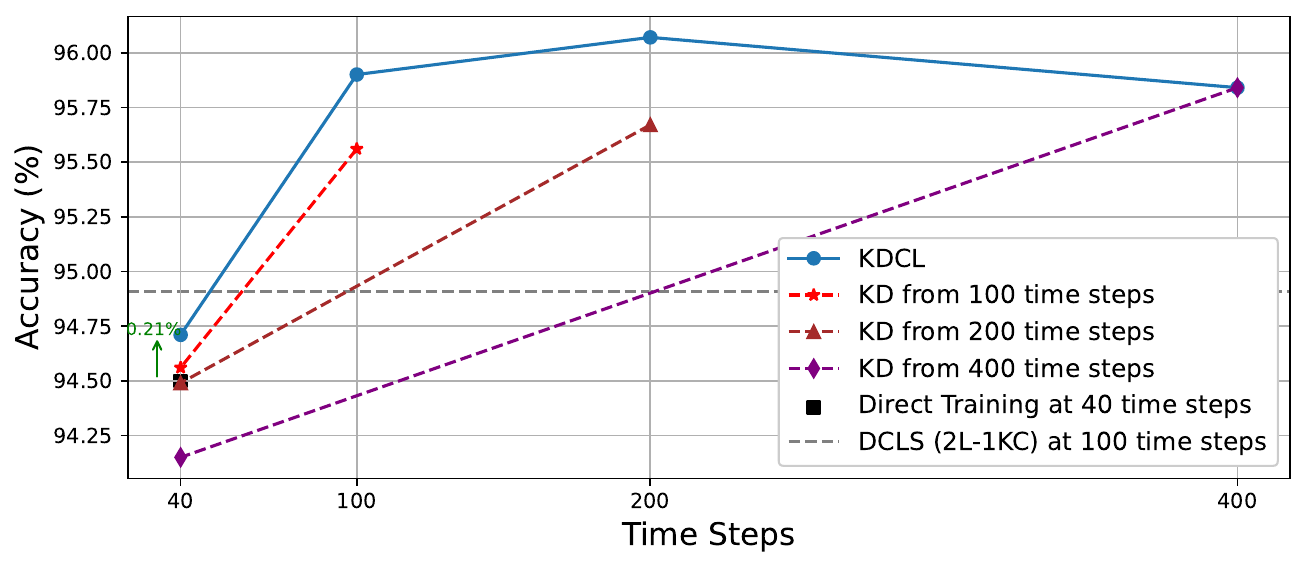}
\end{center}
\caption{Comparing the performance of SpikeSCR with a single block of SGLE among direct training, knowledge distillation, and KDCL on GSC dataset.}\label{fig:kdcl_acc_gsc_1l}
\end{figure}

\begin{figure}[!t]
\begin{center}
\includegraphics[width=1.0\linewidth]{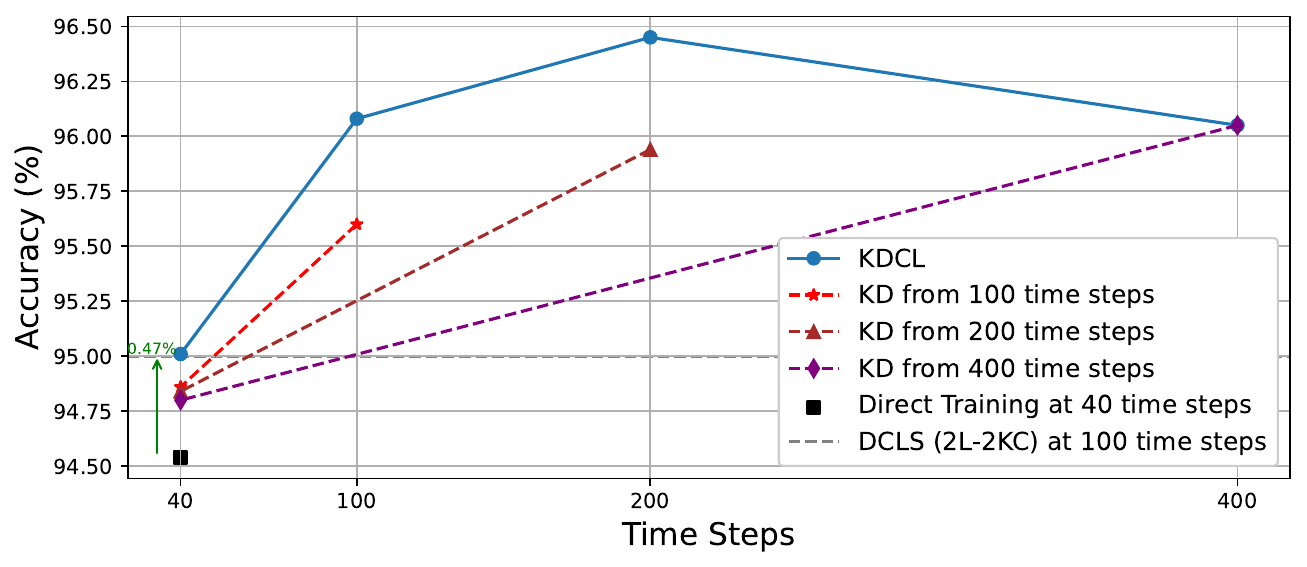}
\end{center}
\caption{Comparing the performance of SpikeSCR with two blocks of SGLE among direct training, knowledge distillation, and KDCL on GSC dataset.}\label{fig:kdcl_acc_gsc_2l}
\end{figure}

\subsection{Appendix C: Comprehensive Results}\label{appendixc}
To provide a more comprehensive understanding of the impact of our SpikeSCR framework and KDCL method across different datasets, we present detailed evaluations in this section, which can showcase the robust adaptability of our proposed framework and method. 

Figure \ref{fig:trade-off compare_ssc} presents a comparative analysis of model performance and energy consumption from short to long time steps on the SSC dataset. We further provide detailed numerical statistics in terms of time steps, parameters, energy and accuracy, showcasing in Table \ref{table:performance_ssc}, and conduct a thorough analysis. Across all time steps, SpikeSCR outperforms the DCLS model, where (3L-2KC) represents a 3-layer feedforward network with dilated convolutions and 2 learnable kernel counts used for learning synaptic delays. With 40 time steps, SpikeSCR exhibits lower energy consumption than DCLS (0.0079mJ and 0.0142mJ vs. 0.0276mJ). With 100 time steps, the energy consumption of SpikeSCR begins to show an increasing trend. Specifically, SpikeSCR with one SGLE block still exhibits competitive energy consumption (0.0171 mJ vs. 0.0216 mJ). However, with two SGLE blocks, its energy consumption surpasses that of DCLS (0.0314 mJ vs. 0.0216 mJ). Moreover, at 200, and 500 time steps, SpikeSCR's energy consumption gradually increases. Although DCLS mitigates rising energy consumption by learning delays with only a few kernel counts, its performance deteriorates with increasing time steps (from 77.77\%, 81.08\%, 80.86\% to 78.76\%), indicating a lack of long-term learning capability. Additionally, we further demonstrate the superior performance of KDCL-based models compared to those trained with the same time steps under only one block of SGLE, as shown in Figure \ref{fig:kdcl_acc_ssc_1l}. Consistent with the conclusions from Figure \ref{fig:kdcl_acc_ssc_2l}, KDCL-based models consistently outperform those trained directly at any specified time step. Direct knowledge distillation from these teacher models with large time steps to student models with only 40 time steps yields inconspicuous outcomes (79.76\% vs. 79.63\%, 79.58\%, and 79.55\%). KDCL achieves the most promising results at 40 time steps, improving accuracy by 0.64\% compared to direct training (79.76\% vs. 79.12\%). The comparable performance with SOTA methods like DCLS (2L-2KC) (79.76\% vs. 80.16\%; 40 steps vs. 100 steps) highlights KDCL's effectiveness with limited time steps.

Figure \ref{fig:trade-off compare_gsc} presents a comparative analysis of model performance and energy consumption across different time steps on the GSC dataset, with detailed numerical statistics provided in Table \ref{table:performance_gsc}. In terms of performance, SpikeSCR continues to demonstrate long-term learning capabilities as the number of time steps increases, showing a consistent upward trend. In contrast, DCLS achieves its best results at 200 time steps but shows a noticeable decline at 400 time steps. For Mel spectrogram inputs, SpikeSCR exhibits greater advantages in energy consumption compared to DCLS. At 40 and 100 time steps, the energy consumption for both one and two SGLE blocks are lower than that of DCLS. Specifically, at 40 time steps, energy consumption is 0.0280 mJ and 0.0384 mJ for one and two SGLE blocks respectively, compared to DCLS's 0.1220 mJ. With 100 time steps, energy consumption is 0.0675 mJ and 0.0939 mJ for SpikeSCR, compared to DCLS's 0.1066 mJ. However, there is a clear upward trend in SpikeSCR's energy consumption at 200 and 400 time steps.

Furthermore, we demonstrate the superior performance of KDCL-based models compared to those trained under the same time steps using one block and two blocks of SGLE on the GSC dataset, as illustrated in Figure \ref{fig:kdcl_acc_gsc_1l} and Figure \ref{fig:kdcl_acc_gsc_2l}. Consistent with the conclusions on the SSC dataset, KDCL-based models outperform those trained directly at any specified time step. Direct knowledge distillation from these teacher models with large time steps to student models with only 40 time steps yields inconspicuous outcomes. For one block of SGLE, it achieves 94.71\% compared to 94.56\%, 94.49\% and 94.15\%. KDCL achieves the most promising results at 40 time steps, improving accuracy by 0.21\% compared to direct training (94.71\% vs. 94.50\%). The comparable performance with DCLS (2L-1KC) (94.71\% vs. 94.91\%; 40 steps vs. 100 steps) highlights its effectiveness with limited time steps. For two blocks of SGLE, it achieves 95.01\% compared to 94.86\%, 94.84\%, and 94.80\%. KDCL achieves the most promising results at 40 time steps, improving accuracy by 0.47\% compared to direct training (95.01\% vs. 94.54\%). The comparable performance with SOTA methods like DCLS (2L-2KC) (95.01\% vs. 95.00\%; 40 steps vs. 100 steps) demonstrates KDCL's efficiency when operating with fewer time steps.

\begin{table}[!t]
\centering
\begin{tabular}{@{}clcc@{}}
\toprule
\textbf{Dataset} & \textbf{Position Encoding} & \textbf{Energy (mJ)} & \textbf{Acc (\%)} \\ \midrule
 \multirow{4}{*}{SSC}
 & None & 0.0173 &  81.79 \\
 & Sinusoidal PE (SinPE)  & 0.0609 & 81.59 \\
 & SinPE + LIF & 0.0145 & 66.48 \\
 & RoPE + LIF & 0.0169 & 81.94 \\ 
\bottomrule
\end{tabular}
\caption{Different positional encodings on SSC dataset.}
\label{fig:compare_pe_ssc}
\end{table}

\begin{table}[!t]
\resizebox{0.5\textwidth}{!}{%
\centering
\begin{tabular}{@{}clcc@{}}
\toprule
\textbf{Dataset} & \textbf{Model} & \textbf{Param (M)} & \textbf{Acc (\%)} \\ \midrule
\multirow{5}{*}{SHD} & SpikeSCR (1L-8-128) & 0.25 & 95.70 \\
                         & w/o SAM & 0.25 & 94.55 \\
                         & w/o RoPE & 0.25 & 93.69 \\
                         & w/o SGU & 0.22 & 92.93 \\
                         & w/o Separable Convolution  & 0.15 & 88.36 \\ 
                         \bottomrule
\end{tabular}
}
\caption{Ablation studies on SHD dataset (100 time steps).}
\label{fig:ablation_shd}
\end{table}

We further discuss the impact of different plug-and-play positional encodings on model performance and energy consumption on the larger-scale SSC dataset, as detailed in Table \ref{fig:compare_pe_ssc}. Compared to the baseline without PE, which achieves an accuracy of 81.79\% and energy consumption of 0.0173 mJ, the SinPE slightly decreases performance to 81.59\% while increasing energy consumption to 0.0609 mJ, about 3.5 times higher, due to the transformation of spike-form data into float-form data. Integrating SinPE with a LIF neuron significantly reduces energy consumption to 0.0145 mJ, but at the cost of dramatically decreasing accuracy to 66.48\%. In contrast, the integration of RoPE with LIF maintains energy consumption at 0.0169 mJ, comparable to the baseline without PE, and achieves the highest accuracy of 81.94\% among the tested encodings.

Comprehensive ablation studies are conducted on the smaller-scale, spike-form SHD dataset with 100 time steps, with results summarized in Table \ref{fig:ablation_shd}.  Eliminating the augment module reduces accuracy by an additional 1.15\%. The impact of removing RoPE from the global representation learning module lead to a 0.86\% reduction in accuracy. Omitting the spiking gated unit from the local representation learning module results in a 0.76\% decrease. Moreover, the absence of the separable convolution module causes a more significant decline, reducing 4.57\%. 


\subsection{Appendix D: Theoretical Calculation of Energy Consumption} \label{appendixd}
The theoretical energy consumption of an SNN is usually calculated through multiplication between the number of MAC/AC operations and the energy consumption of each operation on predefined hardware \cite{panda2020toward,zhou2023spikformer,yao2024spikedriven,zhang2024sglformer}.
The number of synaptic operations (SOPs) is calculated as:
\begin{equation}\label{eq:sop}
    SOP^l=fr^{l-1} \times FLOP^l,
\end{equation}
where $fr^{l-1}$ is the firing rate of spiking neuron layer $l-1$. $FLOP^l$ refers to the number of floating-point MAC operations (FLOPs) of layer $l$, and $SOP^l$ is the number of spike-based AC operations (SOPs).
Assuming the MAC and AC operations are performed on the 45nm hardware \cite{horowitz20141}, i.e. $E_{MAC}=4.6pJ$ and $E_{AC}=0.9pJ$, the energy consumption of SpikeSCR can be calculated as follows:
\begin{equation}\label{eq:energy_SSC}
    E = E_{AC} \times \left(\sum_{i=1}^{N} SOP_{Conv}^i + \sum_{j=1}^M SOP_{SSA}^j\right),
\end{equation}

\begin{equation}\label{eq:energy_GSC}
    E = E_{MAC} \times \left( FLOP_{Conv}^1 \right) + E_{AC} \times \left( \sum_{i=2}^{N} SOP_{Conv}^i + \sum_{j=1}^{M} SOP_{SSA}^j \right)
\end{equation}

$SOP_{Conv}$ represents the SOPs of a convolution or linear layer, and $SOP_{SSA}$ represents the SOPs of an SSA module, $FLOP_{Conv}^1$ represents the FLOPs of the first layer before encoding input frames into spikes. $N$ is the total number of convolution layers and linear layers, and $M$ is the number of SSA modules.  
During model inference, several cascaded linear operation layers such as convolution, linear, and BN layers, can be fused into one single linear operation layer \cite{zhou2024direct}, still enjoying the AC-type operations with a spike-form input tensor.

We further explain the energy consumption calculation under two different input conditions. Specifically, when the input to SpikeSCR is in spike-form (such as the SHD and SSC datasets), it does not involve MAC  operations, and thus its energy consumption is calculated as shown in Eq. \ref{eq:energy_SSC}. However, when the input to SpikeSCR is in real-valued form (such as the GSC dataset), the first Conv layer in the SEM involves MAC operations, and its energy consumption is calculated as shown in Eq. \ref{eq:energy_GSC}. 


\subsection{Appendix E: Code Available}\label{appendixe}
We utilize the SpikingJelly framework \cite{fang2023spikingjelly} to train our model, available at \url{https://github.com/fangwei123456/spikingjelly}. 

For comparison, we select the currently reproducible SOTA work DCLS \cite{hammouamrilearning2024}, with code available at \url{https://github.com/Thvnvtos/SNN-delays}. Moreover, we employ the same data preprocessing methods as those used in the DCLS work to ensure consistency in our experimental setup.

Our energy consumption calculation framework is based on syops-counter \cite{chen2023training}, with code available at \url{https://github.com/iCGY96/syops-counter}.

Our organized code will be made publicly available on a common repository upon reaching the camera-ready version of this paper.

\end{document}